\documentclass[letterpaper, 10 pt, conference, twoside]{ieeeconf}

\IEEEoverridecommandlockouts

\usepackage{amsmath} 
\usepackage{amssymb}  
\usepackage{mathrsfs}
\usepackage{graphicx} 
\usepackage{epsfig} 
\usepackage{xcolor}
\usepackage{bm}
\usepackage{cite}

\newtheorem{theorem}{\textbf{Theorem}}

\newtheorem{definition}{\textbf{Definition}}
\newtheorem{example}{\textbf{Example}}

\newcommand{\Real}{\mathbb R}

\newcommand{\col}{\textrm{col}}

\newcommand{\diag}{\textrm{diag}}

\newcommand{\iden}{\mathbb{I}}

\newcommand{\op}{\textrm{op}}
\newcommand{\des}{\textrm{des}}

\newcommand{\gap}{\vspace{0.1cm}}
\DeclareMathOperator*{\argmin}{argmin}

\begin{document}


\title{\LARGE \bf{Safety-Critical Coordination for Cooperative Legged Locomotion \newline via Control Barrier Functions}}


\author{Jeeseop~Kim$^{1}$, Jaemin~Lee$^{1}$, and Aaron~D.~Ames$^{1}$
\thanks{The work is supported by National Science Foundation (NSF) under the Grant 1924526, 1932091, and 1923239.}
\thanks{$^{1}$J. Kim, J. Lee, and A. D. Ames are with the Department of Mechanical and Civil Engineering, California Institute of Technology, Pasadena, CA 91125, USA, {\tt\small \{jeeseop, jaemin87, ames\}@caltech.edu}}%
}

\maketitle

\begin{abstract}
This paper presents a safety-critical approach to the coordinated control of cooperative robots locomoting in the presence of fixed (holonomic) constraints.  To this end, we leverage control barrier functions (CBFs) to ensure the safe cooperation of the robots while maintaining a desired formation and avoiding obstacles.  The top-level planner generates a set of feasible trajectories, accounting for both kinematic constraints between the robots and physical constraints of the environment. This planner leverages CBFs to ensure safety-critical coordination control, i.e., guarantee safety of the collaborative robots during locomotion.
The middle-level trajectory planner incorporates interconnected single rigid body (SRB) dynamics to generate optimal ground reaction forces (GRFs) to track the safety-ensured trajectories from the top-level planner while addressing the interconnection dynamics between agents.
Distributed low-level controllers generate whole-body motion to follow the prescribed optimal GRFs while ensuring the friction cone condition at each end of the stance legs. The effectiveness of the approach is demonstrated through numerical simulations and experimentally on a pair of quadrupedal robots.

\end{abstract}



\section{Introduction}

\begin{figure}[t!]
\centering
\includegraphics[draft=false, width=\linewidth]{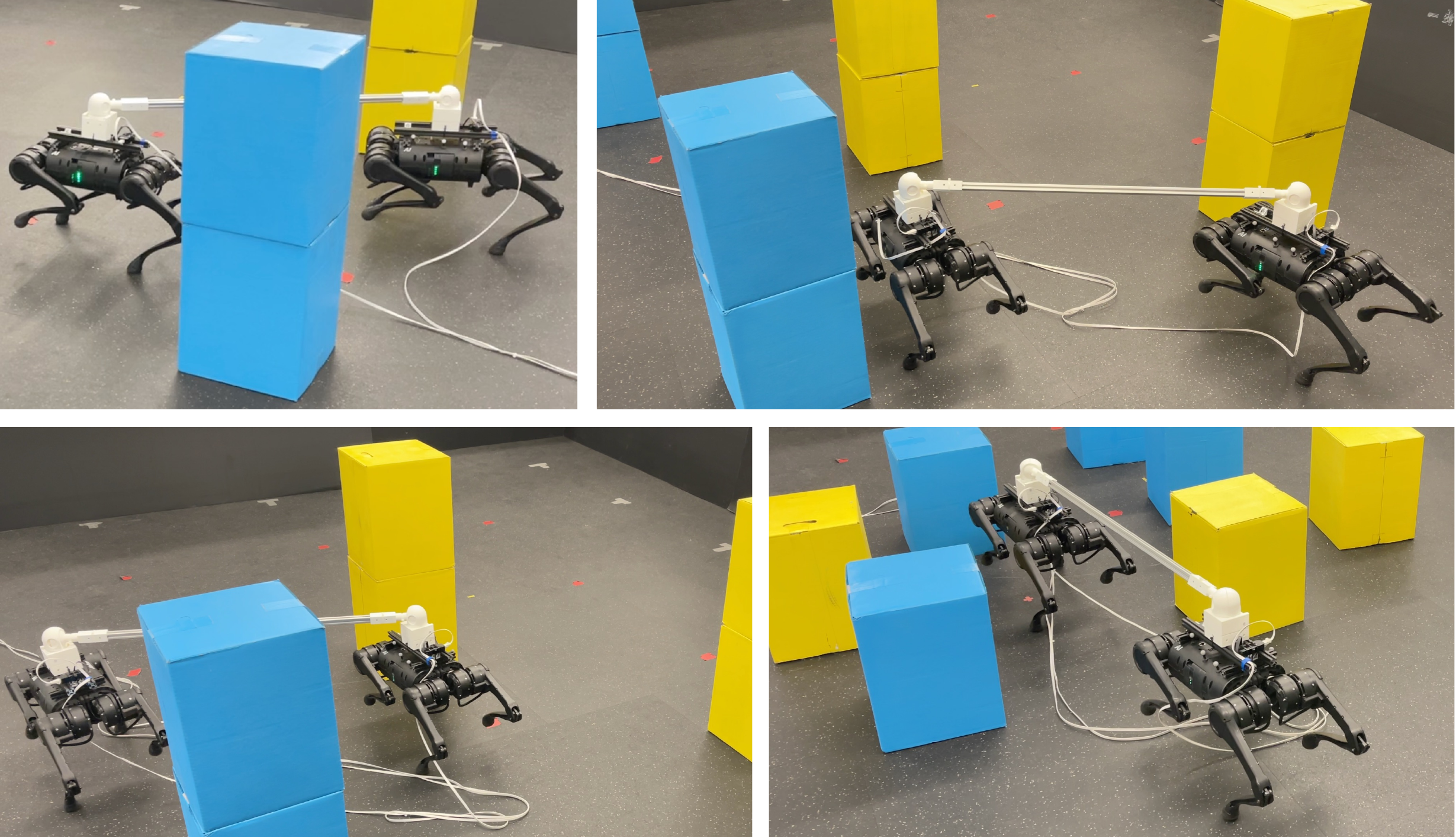}
\vspace{-2.0em}
\caption{Snapshots illustrating the safety-ensured cooperative locomotion of holonomically constrained quadrupedal robots in the environment with obstacles, wherein the CBF-based method avoids collisions.}
\label{fig:titlepic}
\vspace{-1em}
\end{figure}

Collaborative legged robots have the potential to work with each other and to assist people in human-centered environments, e.g., transporting objects in urban settings. One of the essential problems in deploying collaborative legged robots is the mitigation of potentially conflicting objectives: the robots need to maintain a fixed formation (encoded by holonomic constraints) in a cooperative fashion, safely maneuver to avoid obstacles (see Fig. \ref{fig:titlepic}), all while dynamically locomoting. 
While the coordinated control of multi-robot systems (MRSs) subject to holonomic constraints has been studied---notably in the context of manipulators, ground and aerial vehicles \cite{culbertson2021decentralized, machado2016multi, tagliabue2019robust, wehbeh2020distributed}---legged robots present new and unique challenges. 
Collaborative legged robots are highly dynamic, high dimensional, and are hybrid in nature---all of which make synthesizing safety-critical planning and control algorithms more complex. The addition of holonomic constraints between the robots create strong interaction wrenches (forces/torques), which must be carefully addressed to ensure a stable and safe cooperative legged locomotion. 

\subsection{Related Work}

Several studies have introduced control methodologies for stable and robust cooperative locomotion of holonomically constrained legged robots. These frameworks have utilized various techniques, such as the interconnected linear inverted pendulum (LIP) model \cite{kim2022cooperative}, interconnected single rigid body (SRB) dynamics \cite{kim2022layered}, and data-driven trajectory planners \cite{fawcett2022distributed}. However, \textit{safety-critical} collaborative legged locomotion has 
yet to be studied in the the context of holonomically constrained collaborative locomotion. 


Ensuring safety is essential in deploying collaborative robotic systems in environments with obstacles. \textit{Control barrier functions} (CBFs) \cite{ames2016control, ames2019control} are a popular tool for achieving safety guarantees on robotic systems. In particular, CBFs have been successfully applied to achieve collision-free behavior on MRSs\cite{wang2017safety, zhao2017defend, pickem2017robotarium, chen2020guaranteed}. However, designing CBFs and implementing corresponding safe control inputs can be challenging when the system dynamics are complex. To address this issue, reduced-order and model-free approaches have been proposed for single robot systems \cite{squires2021model,molnar2021model,singh2020robust}.
These approaches have demonstrated the effectiveness in leveraging simplified models in the design of CBFs: improving safety without incurring a significant computational burden.
In this context, and in the setting of legged robots, 
recent studies have introduced CBF-based planners/controllers to generate safe foot placement with model predictive control (MPC) via a layered architecture \cite{grandia2021multi}, and ensure safe whole-body motion control \cite{khazoom2022humanoid}. However, addressing the \textit{safety} in the \textit{coordination of the multiple legged robots} in the presence of \emph{holonomic constraints} has not been studied.

\begin{figure}[t!]
\centering
\includegraphics[draft=false, width=\linewidth]{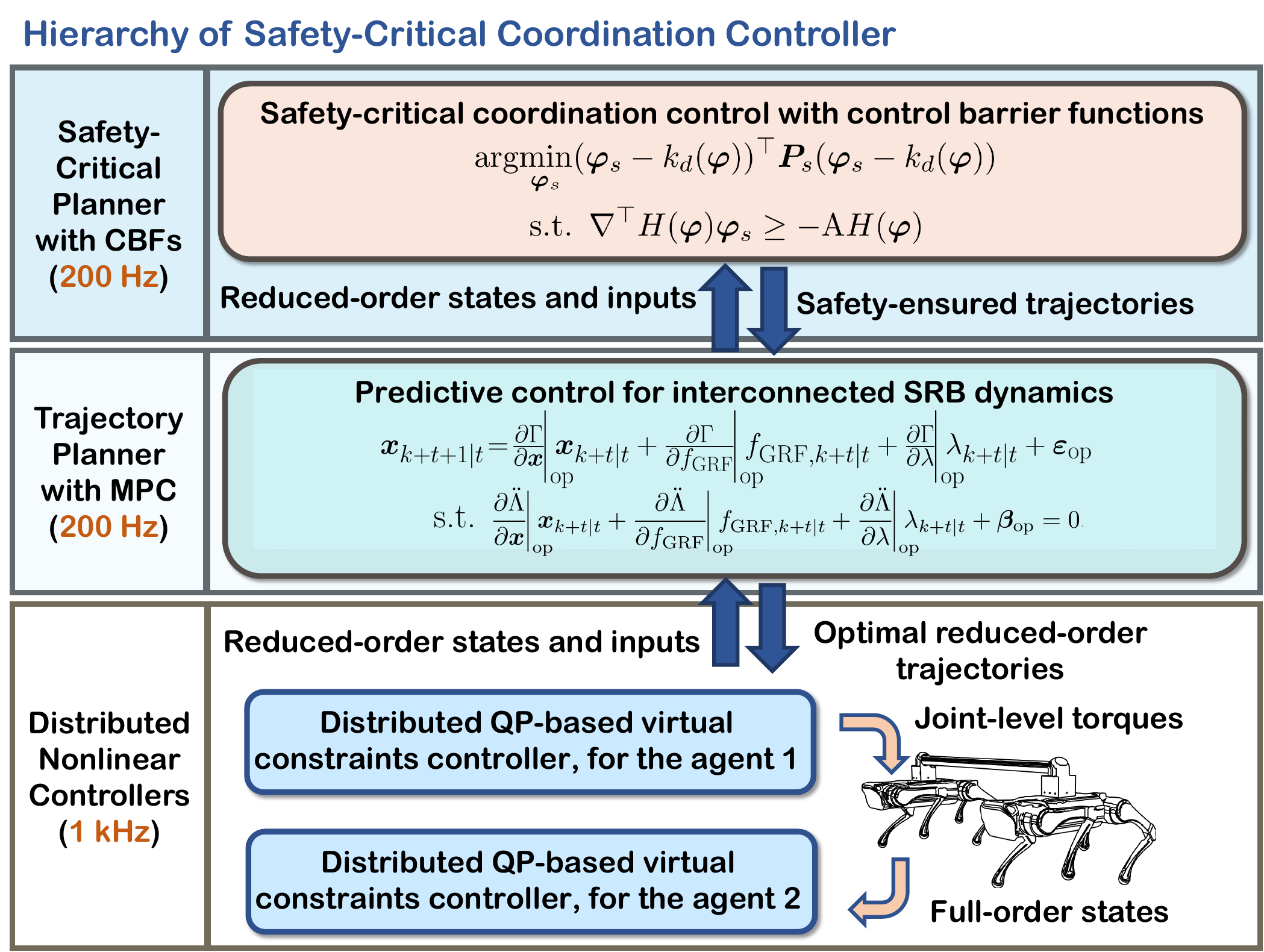}
\vspace{-2.0em}
\caption{Overview of the proposed control approach with safety-critical planner with CBFs at the top level, MPC with interconnected SRB dynamics at the middle level, and the distributed QP-based virtual constraint controllers at the low level for the safe cooperative locomotion.}
\label{fig:overview}
\vspace{-1.8em}
\end{figure}

\subsection{Contributions}
To address the safety-critical control of cooperative legged robots, this paper makes three key contributions: 1) a safety-critical CBF-based planning framework that can guarantee safety in coordinating legged multi-robot systems, 2) a control structure that effectively addresses cooperative locomotion while tracking the safety-ensured trajectories, and 3) experimental evaluation on a pair of quadrupeds (Fig. \ref{fig:titlepic}).


To realize the key contributions of the paper, and thereby ensure safety-critical cooperative locomotion in holonomically constrained legged robots. a three-layered control architecture is proposed (as illustrated in Fig. \ref{fig:overview}):  \newline
\noindent 1) The first layer leverages CBFs to synthesize a safety-critical planner for cooperative locomotion using reduced-order kinematic models of holonomically constrained robots.
\newline 
\noindent 2) The second layer generates optimal ground reaction forces (GRFs), via interconnected SRB dynamics, to track the safety-ensured velocity trajectories from the first layer while considering the dynamics induced by holonomic constraints. \newline
\noindent 3) The third layer generates whole-body motion for each agent while imposing optimal GRFs and safety-ensured velocities from the other two layers. This layer utilizes the full-order model of each agent with I-O linearization of virtual constraints to generate the desired torque inputs. 

The paper is organized as follows. Section \ref{sec:safetyofcooperation} revisits the concept of CBFs and extends it to the setting of safety-critical coordination control. Section \ref{sec:safetyplanner} formulates the safety-critical planner for the kinematics of the interconnected system and formulates the trajectory planner based on interconnected SRB dynamics and MPC, which considers the dynamics of the robot when tracking the safety-ensured velocities.
Section \ref{sec:qpvcctrl} introduces a distributed Quadratic Programming (QP) based virtual constraint controller for whole-body motion control. In Section \ref{sec:validations}, we discuss the numerical and experimental results of the proposed safety-critical coordination control.  
We present conclusions in Section \ref{sec:conclusion}.



\section{Safety of Cooperation}
\label{sec:safetyofcooperation}
In this section, we will examine the safety of cooperative locomotion trajectories for $N$ agents ($N=2$ in our case) using the framework of CBFs. We consider a control-affine system for the $N$ agents:
\begin{equation}\label{eq:controlaffinesystem}
    \dot{\bm{x}} = f(\bm{x}) + g(\bm{x})\bm{u},
\end{equation}
with state $\bm{x}(t)\in \prod_{i \in \mathcal{I}}\mathcal{X}_i$
and control input $\bm{u}(t)\in \prod_{i \in \mathcal{I}} \mathcal{U}_i$.
Here $i\in \mathcal{I}:=\{1, \ldots, N\}$ represents the indexing for each individual agent, wherein the state agent $i$ is $\bm{x}_i \in \mathcal{X}_i$, and the control input of the agent $i$ is $\bm{u}_i \in \mathcal{U}_i$.
We assume $f: \mathcal{X} \rightarrow \mathbb{R}^{n}$ and $g: \mathcal{X}\rightarrow\mathbb{R}^{n\times m}$ are Lipschitz continuous, where $n=\sum_{i\in\mathcal{I}}n_i$ and $m=\sum_{i\in\mathcal{I}}m_i$.

\subsection{Control Barrier Function and Safety of Individual Agent}
We consider \eqref{eq:controlaffinesystem} safe if its state $\bm{x}(t)$ is located within a \textit{safe set} $\mathcal{S} \subset \mathcal{X}$ for all time, i.e., if $\mathcal{S}$ is forward invariant: 

\gap
\begin{definition}\label{def:safety}
System \eqref{eq:controlaffinesystem} is \textit{safe} with respect to the set $\mathcal{S}$ if the set $\mathcal{S}$ is \textit{forward invariant}: if for every $\bm{x}_0 \in \mathcal{S}$, $x(t)\in\mathcal{S}$ for every $\bm{x}(0)=\bm{x}_0$ and $\forall t \geq 0$.
\end{definition}
\gap

The safe set is determined based on the system configuration. For example, it can include robot positions that prevent collisions with nearby obstacles, or it can reflect the positional relationship between robots to describe the holonomic constraint. Specifically, we define the safe set $\mathcal{S}$ as the \textit{superlevel set} of a continuously differentiable function $h:\mathcal{X}\rightarrow \mathbb{R}$:
\begin{eqnarray}\label{eq:safeset}
    \mathcal{S} & := & \{ \bm{x} \in \mathcal{X} : h(\bm{x}) \geq 0 \}, \\
    \partial \mathcal{S} & := & \{ \bm{x} \in \mathcal{X} : h(\bm{x}) = 0 \}. 
\end{eqnarray}
CBFs can then be utilized as a tool to synthesize controllers that are provably safe, similar to how Control Lyapunov Functions (CLFs)\cite{sontag1999control} ensure stability.

\gap
\begin{definition}
    A continuously differentiable function $h:\mathcal{X}\rightarrow\mathbb{R}$ is a \textit{control barrier function (CBF)} for \eqref{eq:controlaffinesystem} if there exists an extended class $\mathcal{K}_\infty$ function $\alpha \in \mathcal{K}^e_\infty$ such that:
    \begin{equation}\label{eq:cbfdefinition}
        \sup_{u\in\mathcal{U}} \,\,\,\, \dot{h}(\bm{x},\bm{u}) \geq -\alpha(h(\bm{x})),
    \end{equation}
    for all $\bm{x} \in \mathcal{S}$, where 
    $$
    \dot{h}(\bm{x},\bm{u}) = \underbrace{\frac{\partial h(\bm{x})}{\partial \bm{x}}f(\bm{x})}_{:= L_f h(\bm{x})} + 
    \underbrace{\frac{\partial h(\bm{x})}{\partial \bm{x}}g(\bm{x})}_{:=L_g h(\bm{x})}
    $$
    with $L_f h(\bm{x})$ and $L_g h(\bm{x})$ the Lie derivatives \cite{khalil2002nonlinear} of $h$ with respect to $f(\bm{x})$ and $g(\bm{x})$, respectively. 
\end{definition}
\gap 

Note that for simplicity we often pick $\alpha(h(\bm{x}))$ to be $\alpha h(\bm{x})$, where $\alpha \in \mathbb{R}_{>0}$. 

\gap
\begin{theorem}[\hspace{-0.01cm}\cite{ames2016control}]\label{th:cbf}
\emph{If $h$ is a \textit{CBF} for \eqref{eq:controlaffinesystem}, then any locally Lipschitz continuous controller $\bm{u}=k(\bm{x})$ satisfying
\begin{equation}\label{eq:cbfsafecondition}
    \dot{h}(\bm{x}, k(\bm{x})) \geq -\alpha(h(\bm{x})), \qquad \forall \bm{x}\in \mathcal{S}
\end{equation}
guarantees that \eqref{eq:controlaffinesystem} is safe with respect to $\mathcal{S}$.}
\end{theorem}
\gap

Ensuring the safety of each agent can be easily accomplished by solving a QP. The objective is to minimize the difference between the desired control input and the actual input for each agent, subject to an inequality constraint that ensures the agent avoids collisions with an obstacle. The constraint for safety is represented by the function $h_i(\bm{x})$, which represents the CBF for the $i$th agent with respect to the obstacle. This can be expressed, for $i\in\mathcal{I}$, as the QP:
\begin{alignat}{4}\label{eq:qpforoneagent}
    k_i(\bm{x})& = &  \argmin_{\bm{u}_i \in \mathcal{U}_i}&(\bm{u}_i-\bm{u}_i^d)^\top \bm{P}_s (\bm{u}_i-\bm{u}_i^d)  \\
    &&\mathrm{s.t.} & \,\,\, L_f h_i(\bm{x}) + L_g h_i(\bm{x}) \bm{u}_i \geq -\alpha h_i(\bm{x}), \nonumber
\end{alignat}
where $\bm{u}_i^d$ is the desired control input for the $i$th agent, $\bm{P}_s$ is a positive definite matrix, and $\alpha$ is a positive number that can be tuned. The result of the QP is $k_i(\bm{x})$,  the (pointwise) optimal control input that meets the CBF constraint to ensure the safety of the $i$th agent.  The constraint on the decision variable, $\bm{u}_i \in \mathcal{U}_i$, implies that the optimal control input needs to take values in the admissible control input set.  Note that the number of inequality constraints in \eqref{eq:qpforoneagent} can be increased based on the number of obstacles that need to be avoided by the $i$th agent. More specifically, if there are $D$ number of obstacles to be avoided, inequality constraints in \eqref{eq:qpforoneagent} can be written as $L_f h^{z}_i(\bm{x}) + L_g h^{z}_i(\bm{x}) \bm{u}_i \geq -\alpha h^{z}_i(\bm{x})$ where $z\in\mathcal{Z}:=\{1,\dots,D\}$ and $h^{z}_i(\bm{x})$ represents the CBF for the $i$th agent with respect to the $z$th obstacle.

The QP given in \eqref{eq:qpforoneagent} is able to address the safety of each individual agent in relation to the obstacles in their environment. However, when multiple agents are working together and subject to holonomic constraints, it becomes necessary to consider \textit{safe coordination} while also \textit{adhering to the holonomic constraint} for safe cooperative locomotion.

\subsection{Control Barrier Function for Coordination}
When coordinating cooperative locomotion, it is crucial to consider the safety of both the individual agents themselves and the space between the agents, particularly when holonomic constraints or payload are involved. For example, during an object transportation task by two robots, both robots must avoid collision and ensure that the object is also safely transported without any collisions.

The function representing the positions where the holonomic constraint is imposed on each agent and the spatial positions between them is given by:
\begin{equation}\label{eq:internalpos}
    \rho_{ij}(\bm{x}) = \gamma p_i(\bm{x}) + (1-\gamma) p_j(\bm{x}),
\end{equation}
where $\gamma \in \Upsilon=\{\gamma \in \mathbb{Q}: 0\leq \gamma \leq 1\}$ is a weighting factor, $p_i(\bm{x})$ and $p_j(\bm{x})$ represents the constraint applied position on the $i$th and $j$th agent, respectively (with $i,j\in\mathcal{I}$ and $i\neq j$). Here we note that $\rho_{ij}(\bm{x})$ represents the variation in space between $p_i(\bm{x})$ and $p_j(\bm{x})$ as $\gamma$ is varied. For ease of use, we establish the bijective function $c:\Upsilon\rightarrow \{1,\dots,\eta\}$ where $\eta$ represents the total number of distinct spatial positions between the agents.

Avoiding the obstacle vai coordination control can be achieved through the use of a CBF leveraging $\rho_{ij}(\bm{x})$:
\begin{equation}\label{eq:cbfofcoordination}
    h^{c(\gamma)}_{\textnormal{co}}(\bm{x}) = \lVert \rho_{ij}(\bm{x}) - \bm{q}\rVert - r,
\end{equation}
where $r\in \mathbb{R}_{>0}$ is an obstacle radius and $\bm{q}$ is the center position of the obstacle. Here we remark that $h^{c(\gamma)}_{\textnormal{co}}(\bm{x})$ represents the CBF for a specific spatial position depending on $\gamma$ along the holonomic constraint. The total number of such positions is $\eta$. Furthermore, the obstacle specification in \eqref{eq:cbfofcoordination} can be represented with a tuple $(\bm{q}, r)$. To ensure safety in environments containing multiple obstacles, multiple CBFs can be generated by using different tuples to address each obstacle individually. Thus the CBFs for safe coordination control with respect to multiple obstacles can be written as $h^{c(\gamma), z}_{\textnormal{co}}(\bm{x})$, where $z$ represents the $z$th obstacle.

Analogus to \eqref{eq:qpforoneagent}, a QP with the constraints for safety represented by the CBFs for each agent and CBFs for coordination can be written as:
\begin{alignat}{4}\label{eq:qpforagents}
    k(\bm{x}) & = &\argmin_{\bm{u} \in \mathcal{U}}&(\bm{u}-\bm{u}^d)^\top \bm{P}_s (\bm{u}-\bm{u}^d) \nonumber \\
    &&\mathrm{s.t.} & \,\,\, L_f H(\bm{x}) + L_g H(\bm{x}) \bm{u} \geq -\mathrm{A} H(\bm{x}),
\end{alignat}
where $\bm{u}^d$ is the desired control input for the cooperative locomotion, $\bm{P}_s$ is a positive definite matrix, $H(\bm{x}) = \col(h^{z}_i(\bm{x}), h^{c(\gamma),z}_{\textnormal{co}}(\bm{x})) \in \mathbb{R}^{D(N+\eta)}$ where $i\in\mathcal{I}$, and $\mathrm{A} = \diag(\alpha^{z}_i, \alpha^{c(\gamma),z}_{\textnormal{co}})\in \mathbb{R}^{D(N+\eta)\times D(N+\eta)}$ is a positive definite diagonal matrix. Here ``$\col$'' and ``$\diag$'' denotes the column operator and vector-to-diagonal matrix operator, respectively. Additionally, $\alpha^{c(\gamma),z}_{\textnormal{co}}$ is the positive constant defining the class $\mathcal{K}_{\infty}^e$ function for the CBFs in \eqref{eq:cbfofcoordination}. We note that the dimension of the $H(\bm{x})$ and $\mathrm{A}$ can vary based on $D$ and $\eta$ that define the total number of obstacles and distinct spacial positions, respectively. 

\gap 

\begin{example}
Let's consider a scenario where we need to ensure safety-critical coordination for two agents in an environment with one obstacle. To guarantee a clear path for the object that the agents aim to transport, additional CBFs are considered on four different positions between agents. To address the safety-critical coordination control in this scenario, $H(\bm{x})$ and $\mathrm{A}$ in \eqref{eq:qpforagents} become $\col(h^1_1(\bm{x}), h^1_2(\bm{x}), h^{1,1}_{\textnormal{co}}(\bm{x}),  h^{2,1}_{\textnormal{co}}(\bm{x}),  h^{3,1}_{\textnormal{co}}(\bm{x}),  h^{4,1}_{\textnormal{co}}(\bm{x}))$ and $\diag(\alpha^1_1,\alpha^1_2, \alpha^{1,1}_{\textnormal{co}},\alpha^{2,1}_{\textnormal{co}},\alpha^{3,1}_{\textnormal{co}},\alpha^{4,1}_{\textnormal{co}})$, respectively.
\end{example}

\gap 

To ensure safe cooperative locomotion in environments with obstacles, the optimization problem formulated as a QP in \eqref{eq:qpforagents} with CBFs in \eqref{eq:cbfofcoordination} can be used. 
However, while the safety-critical coordination control in \eqref{eq:qpforagents} can ensure safe and coordinated cooperative locomotion, it does not force agents to respect the holonomic constraints between them. Therefore, it is crucial to implement additional CBFs that are specifically designed to address these holonomic constraints.

\subsection{Control Barrier Function for Holonomic Constraint}
We addressed the safety of individual agents and the safety of cooperative locomotion with coordination control by formulating QP with CBFs. However, the introduced coordination control for safety in \eqref{eq:qpforagents} still does not consider the relative distance constraint in cooperative locomotion. More specifically, \eqref{eq:qpforagents} can generate optimal control inputs to ensure the coordinated safety of individual agents, but the distance between agents can be changed freely. To address the constant distance constraint case, this section develops CBFs from holonomic constraints.

To ensure the safety of cooperative locomotion subject to holonomic constraints between agents, we first formulate the holonomic constraint using the states of each agent. This can be expressed as:
\begin{equation}\label{eq:nominalholocon}
    \lVert p_i(\bm{x}) - p_j(\bm{x})\rVert = \sqrt{\psi},
\end{equation}
where $\lVert \cdot \rVert$ denotes the 2-norm, $i\in\mathcal{I}$, $j \neq i \in \mathcal{I}$, $p_i(\bm{x})$ and $p_j(\bm{x})$ represents the position on the $i$th and $j$th agent, respectively, on which the holonomic constraints are imposed, and $\psi\in\mathbb{R}_{>0}$ is the constant value that represents the square of the length of the holonomic constraint.


In order to relate the holonomic constraint with the formulation of a safe set given in Definition \ref{def:safety}, \eqref{eq:nominalholocon} can be used in to synthesize two CBFs as follows:
\begin{alignat}{2}\label{eq:cbfforholocon}
    h^1_{\textnormal{hc}}(\bm{x})& = \lVert p_i(\bm{x}) - p_j(\bm{x})\rVert^2 - (1-\epsilon)\psi \nonumber\\
    h^2_{\textnormal{hc}}(\bm{x})& = (1+\epsilon)\psi - \lVert p_i(\bm{x}) - p_j(\bm{x})\rVert^2,
\end{alignat}
where $\epsilon \in \mathbb{R}_{>0}$ is the relaxation variable. Here we note that \eqref{eq:cbfforholocon} with the definition of a safe set in \eqref{eq:safeset} is identical with \eqref{eq:nominalholocon} when $\epsilon$ becomes zero. We will examine the system safety with respect to the tracking error on the control input in Section \ref{sec:issf}, utilizing the $\epsilon$ introduced in \eqref{eq:cbfforholocon}.
The safety-critical coordination control can incorporate the holonomic constraint by using two CBFs, which can be achieved by replacing the $H(\bm{x})$ and $\mathrm{A}$ matrix in \eqref{eq:qpforagents} with:
\begin{align}
    H(\bm{x}) & =\col(h^{z}_i(\bm{x}), h^{c(\gamma),z}_{\textnormal{co}}(\bm{x}), h^1_{\textnormal{hc}}(\bm{x}),h^2_{\textnormal{hc}}(\bm{x})) \nonumber\\
\mathrm{A} & =\diag(\alpha^z_i,\alpha^{c(\gamma),z}_{\textnormal{co}},\alpha^1_{\textnormal{hc}},\alpha^2_{\textnormal{hc}}) \nonumber
\end{align}
with $H(\bm{x}) \in \mathbb{R}^{D(N+\eta)+2}$ and $\mathrm{A} \in \mathbb{R}^{D(N+\eta)+2 \times D(N+\eta)+2}$. 
Here, $i\in\mathcal{I}$ and $\mathrm{A}$ is a positive definite diagonal matrix. Additionally, $\alpha^{1}_{\textnormal{hc}}$ and $\alpha^{2}_{\textnormal{hc}}$ are the linear class $\mathcal{K}_{\infty}^e$ function for the CBFs $h^1_{\textnormal{hc}}(\bm{x})$ and $h^2_{\textnormal{hc}}(\bm{x})$ defined in \eqref{eq:cbfforholocon}, respectively. Similar to \eqref{eq:qpforagents}, the dimension of the $H(\bm{x})$ and $\mathrm{A}$ can vary based on the total number of obstacles, $D$, and $\eta$ that defines the total number of spacial positions determined by $\gamma$ in \eqref{eq:cbfofcoordination}. 


\subsection{Input to State Safety of the Holonomic Constraint}\label{sec:issf}
In practice, coordination control inputs are often subject to unknown disturbances that may affect on safety. To ensure safety in the presence of bounded perturbation, $\xi$, safety in Definition \ref{def:safety} can be recast as\textit{input-to-state safety (ISSf)}\cite{kolathaya2018input} if the system stays within an open set $\mathcal{S}_d \supseteq \mathcal{S}$ that depends on the magnitude of the perturbation: $\| \xi \|_{\infty}$. Here the open set, $\mathcal{S}_d$, is given by:
\begin{equation}\label{eq:issfsafeset}
    \mathcal{S}_d = \{ \bm{x}\in\mathcal{X}: h(\bm{x}) + \kappa(\lVert \xi \rVert_{\infty}) \geq 0 \},
\end{equation}
for some class $\mathcal{K}$ function $\kappa$. It was established in \cite{kolathaya2018input} that ISSf is ensured by the existence of any locally Lipshitz continuous controller $\bm{u}=k(\bm{x})$ satisfying
\begin{equation}\label{eq:issfcbf}
    \dot{h}(\bm{x}, k(\bm{x})) \geq -\alpha(h(\bm{x})) - \iota(\lVert \xi \rVert_{\infty}),
\end{equation}
for some class $\mathcal{K}$ function $\iota$ and $\alpha \in \mathcal{K}_{\infty}^e$.

To ensure safe cooperative locomotion while adhering to the holonomic constraint, we proposed a coordination controller that incorporates holonomic constraints using the two CBFs in \eqref{eq:cbfforholocon}. The relaxation variable $\epsilon$ in \eqref{eq:cbfforholocon} can be rearranged as:
\begin{alignat}{2}\label{eq:issfcbfforholocon}
    h^1_{\textnormal{hc}}(\bm{x})& = \lVert p_i(\bm{x}) - p_j(\bm{x})\rVert^2 - \psi +\psi\epsilon \nonumber\\
    h^2_{\textnormal{hc}}(\bm{x})& = \psi - \lVert p_i(\bm{x}) - p_j(\bm{x})\rVert^2 +\psi\epsilon.
\end{alignat}
Here we note that \eqref{eq:issfcbfforholocon} is in the form of the relaxed safe set for ISSf as given in\eqref{eq:issfsafeset}. More specifically, finding the controller that satisfies the condition:
\begin{alignat}{2}\label{eq:issfcbfineq}
\dot{h}^\sigma_{\textnormal{hc}}(\bm{x}, \bm{u})& \geq -\alpha^\sigma_{\textnormal{hc}}h^\sigma_{\textnormal{hc}}-\iota(\epsilon), \qquad 
\sigma \in \{1,2\}
\end{alignat}
with $h^\sigma_{\textnormal{hc}}$ in \eqref{eq:issfcbfforholocon}, ensures safety with respect to perturbations on the control input with maximum magnitude identical with $\epsilon$. 
The further examination of ISSf in the context of safety-critical coordination can be found in Section \ref{sec:validations}.


\section{Safety Critical Multi-Layered Trajectory Controller for Cooperative Locomotion}\label{sec:safetyplanner}
The objective of this section is to present the multi-layered safety-critical trajectory planner for two holonomically constrained quadrupedal robots. Here we note that the framework introduced in Section \ref{sec:safetyofcooperation}
is applicable to a wide range of robotic systems, yet here we specialize it to the cooperative locomotion of holonomically constrained quadrupedal robots ($\mathcal{I} = \{1,2\}$, as shown in Fig. \ref{fig:titlepic}). To address the safety-critical cooperative control of these qudrupedal robots, we propose a multi-layered trajectory planner comprised of three levels: (top layer) a safety-critical planner level with the kinematics of the interconnection, (middle layer) an optimal GRFs trajectory planner based on MPC with interconnected SRB dynamics, and (bottom layer) a real-time control level for the nonlinear dynamics leveraging virtual constraints.
This three-level approach enables the efficient and effective tracking of safe trajectories when considering the kinematics and dynamics of the holonomically constrained robotic system.

\begin{figure}[t!]
\centering
\includegraphics[draft=false, width=\linewidth]{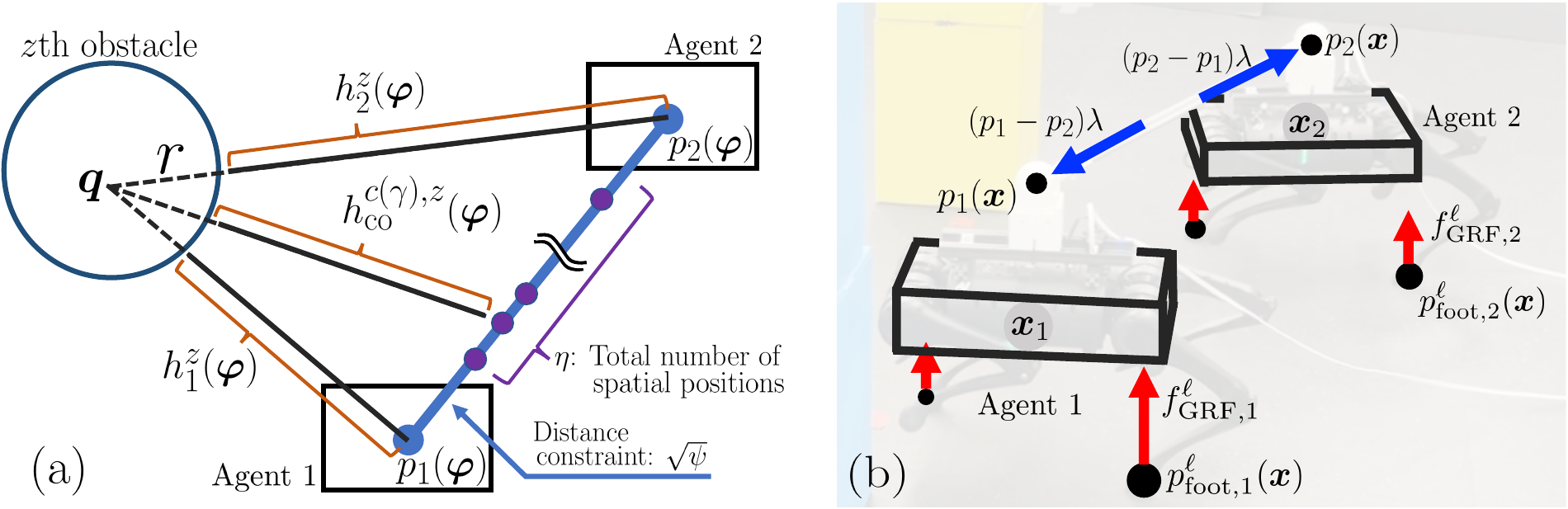}
\vspace{-2.0em}
\caption{Illustration of the holonomically constrained reduced order models  for (a) safety-critical planner with CBFs and (b) trajectory planner with MPC.}
\label{fig:interconnectedrom}
\vspace{-1.5em}
\end{figure}

\subsection{Planner with Safety-Critical Coordination Control}
The safety-critical planner employs holonomically constrained reduced-order models that capture the kinematic properties of the system to guarantee the planned trajectory satisfies the safety constraints. The system of the simple reduced-order model can be written as:
\begin{equation}\label{eq:unitmassmodel}
    \begin{bmatrix}
        \dot{\bm{\varphi}}\\
        \ddot{\bm{\varphi}}
    \end{bmatrix} = 
    \begin{bmatrix}
        0 & \mathbb{I} \\ 0 & 0
    \end{bmatrix} \begin{bmatrix} \bm{\varphi} \\ \dot{\bm{\varphi}} \end{bmatrix} +
    \begin{bmatrix} 0\\ \mathbb{I} \end{bmatrix}\bm{\vartheta},
\end{equation}
where $\mathbb{I}$ represents identity matrix, $\bm{\vartheta}\in\mathbb{R}^4$ denotes the input of double-integrator system, and $\bm{\varphi}=\col(\varphi_1, \varphi_2)\in\mathbb{R}^4$ represents the planar positions of the agents. Here, $\varphi_i\in\mathbb{R}^2$ is the planar position of the agent $i$ ($i\in\mathcal{I}$). Additionally, the holonomic constraint can be written as $\lVert p_1(\bm{\varphi}) - p_2(\bm{\varphi})\rVert = \sqrt{\psi}$, which is described in Fig. \ref{fig:interconnectedrom} (a). 
Analogous to Section \ref{sec:safetyofcooperation}, safety-critical coordination control can be expressed as: 
\begin{alignat}{4}\label{eq:qpforagentstotal}
    &&\arg\!\min_{\bm{\varphi}_s}&(\bm{\varphi}_s-k_d(\bm \varphi))^\top \bm{P}_s (\bm{\varphi}_s-k_d(\bm \varphi)) \nonumber \\
    &&\mathrm{s.t.} & \,\,\, \nabla^{\top} H(\bm{\varphi}) \bm{\varphi}_s\geq -\mathrm{A} H(\bm{\varphi})
\end{alignat}
where $\bm{P}_s$ is a positive definite matrix and $k_d(\bm \varphi)$ is a nominal desired velocity without the safety consideration. Additionally, $\bm{\varphi}_s$ is the safe velocity with the consideration of CBFs. 
Here we note that $H(\bm{\varphi})\in \mathbb{R}^{D(N+\eta)+2}$ is composed of 1) CBFs for each agent with respect to the $z$th obstacle, $h^{z}_i(\bm{\varphi})$, 2) CBFs for the $c(\gamma)$th position between agents with respect to $z$th obstacle, $h^{c(\gamma),z}_{\textnormal{co}}(\bm{\varphi})$, and 3) CBFs for representing holonomic constraints, $h^1_{\textnormal{hc}}(\bm{\varphi}),h^2_{\textnormal{hc}}(\bm{\varphi})$. Similarly, $\mathrm{A}=\diag(\alpha^z_i,\alpha^{c(\gamma),z}_{\textnormal{co}},\alpha^1_{\textnormal{hc}},\alpha^2_{\textnormal{hc}})\in \mathbb{R}^{(DN+D\eta+2) \times (DN+D\eta+2)}$ is composed of 1) $\alpha^z_i$ for each agent with respect to $z$th obstacle, 2) $\alpha^{c(\gamma),z}_{\textnormal{co}}$ for the $c(\gamma)$th position between agents with respect to $z$th obstacle, and 3) $\alpha^1_{\textnormal{hc}},\alpha^2_{\textnormal{hc}}$ for the holonomic constraints. An illustration of the CBFs in the safety-critical planner can be found in Fig. \ref{fig:interconnectedrom} (a).

\subsection{Trajectory Planner with Interconnected SRB Dynamics}
The middle-level trajectory planner focuses on achieving the safe cooperative locomotion by bridging the gap between the safe trajectories generated by the safety-critical planner and the low-level whole-body motion controller which utilizes the nonlinear full-order dynamics. To achieve this, interconnected SRB dynamics are used in the middle-level trajectory planner. The interconnected SRB dynamics enable the computation of optimal GRFs that impose the full-order system to follow the safety-ensured trajectories while accounting for the interconnection dynamics. 

In the notation of interconnected SRB dynamics, the Cartesian coordinates of the center of mass (COM) of agent $i$ in the inertial frame, $\{O\}$, are written as $\phi_{\textnormal{COM}, i}\in \mathbb{R}^3$. The orientation of the body frame, $\{B_i\}$, of agent $i$ with respect to the inertial frame, $\{O\}$ is represented by $R_i \in  \textnormal{SO}(3)$ and $\omega_i^{B_i}\in\mathbb{R}^3$ denotes the angular velocity of agent $i$ in the body frame, $\{B_i\}$. 
Analogous to the variational-based approach of \cite{ding2021representation}, the state variables of the agent $i\in\mathcal{I}$ can be described as $\bm{x}_i:=\col(\phi_{\textnormal{COM}, i}, \dot{\phi}_{\textnormal{COM},i}, \xi_i, \omega_i^{B_i}) \in \mathbb{R}^{12}$
where $\xi_i\in\mathbb{R}^3$ is a vector that approximate the rotation matrix $R_i$ around the operation point $R^{\op}_i$ using the Taylor series expansion as
\begin{equation}\label{eq:rotationmtxTaylor}
    R_i = R^{\op}_i \textnormal{exp}(\begin{bmatrix}\xi_i\end{bmatrix}_{\times}) \approx R^{\op}_i (\mathbb{I} + \begin{bmatrix}\xi_i\end{bmatrix}_{\times}),
\end{equation}
where the map $\begin{bmatrix} \cdot \end{bmatrix}_{\times}: \mathbb{R}^3 \rightarrow \mathfrak{so}(3)$ represents the skew-symmetric operator. The approximation of the rotation matrix, $R_i$ in \eqref{eq:rotationmtxTaylor} helps to evolve approximately on $\textnormal{SO}(3)$. 

The holonomic constraint in interconnected SRB dynamics can be described as $\Lambda(\bm{x}):=\lVert p_i(\bm{x})-p_j(\bm{x})\rVert = \sqrt{\psi}$
 where $p_i(\bm{x})$ and $p_j(\bm{x})$ represents the position of the holonomic constraint end on agent $i$ and $j$, respectively, which is described in Fig. \ref{fig:interconnectedrom} (b). Here we note that the holonomic constraint is the constraint on the Euclidean distance between $p_i(\bm{x})$ and $p_j(\bm{x})$. According to the principle of virtual work, $(p_i - p_j)\lambda \in \mathbb{R}^3$ can be considered to be the interaction force applied on agent $i$ for some Lagrange multiplier $\lambda\in\mathbb{R}$. Accordingly, the net force, $f_i^{\textnormal{net}}$, and net torque ,$\tau_i^{\textnormal{net}}$, applied to agent $i$ are described by:
\begin{alignat}{2}\label{eq:wrench}
    \begin{bmatrix}
        f_i^{\textnormal{net}}\\ \tau_i^{\textnormal{net}}
    \end{bmatrix} =
    \sum_{\ell}&\begin{bmatrix} \mathbb{I}\\ \begin{bmatrix}p^{\ell}_{\textnormal{foot},i}(\bm{x}) - \bm{x}_i\end{bmatrix}_{\times}\end{bmatrix}f_{\textnormal{GRF},i}^{\ell} \nonumber \\
    &+\begin{bmatrix} \mathbb{I}\\ \begin{bmatrix}p_{i}(\bm{x}) - \bm{x}_i\end{bmatrix}_{\times}\end{bmatrix}(p_i-p_j)\lambda,
\end{alignat}
where $j\in\mathcal{I}$ denotes the index of the other agent and the superscript $\ell \in \mathcal{L}_i=\{1,2,3,4\}$ denotes the index of the contacting legs on the ground for the agent $i$. Finally, the inerconnected SRB dynamics can be expressed as 
\begin{equation}\label{eq:interconnectedSRBdyn}
    \dot{\bm{x}} = \Gamma(\bm{x}, f_{\textnormal{GRF}}, \lambda)\,\,\, \textnormal{s.t}\,\, \ddot{\Lambda}(\bm{x}, f_{\textnormal{GRF}}, \lambda) =0,
\end{equation} where $\bm{x}=\col(\bm{x}_1, \bm{x}_2)\in\mathcal{X}$ with \eqref{eq:wrench}, $\dot{R}_i = R_i\begin{bmatrix}\omega_i^{B_i} \end{bmatrix}_{\times}$, and $I\dot{\omega_i}^{B_i}+\begin{bmatrix} \omega_i^{B_i} \end{bmatrix}_{\times}I\omega_i^{B_i}=R_i^{\top}\tau_i^{\textnormal{net}}$. Additionally, $I\in \mathbb{R}^{3\times 3}$ denote the moment of inertia of each agent with respect to its body frame. 

We note that the variational-based linearization in \cite{chignoli2020variational, ding2021representation} has linearized SRB dynamics subject to GRFs without interaction forces, achieved by linearizing rotational portions with \eqref{eq:rotationmtxTaylor}. However, the interaction forces induced by the interconnection between SRB models still impose the dynamics of a nonlinear system \eqref{eq:interconnectedSRBdyn}, although \eqref{eq:rotationmtxTaylor} linearizes the rotational portions of the interconnected SRB dynamics. Similar to \cite{kim2022cooperative}, this can further be linearized and discretized using Jacobian linearization and forward Euler method, respectively.   

With the goal of utilizing MPC with this linearized and discretized system with $N_{\textnormal{h}}$ control horizon, a discrete and linear time-varying (LTV) system can be written as follows:
\begin{equation}\label{eq:linearizedsrb}
    \bm{x}_{k+t+1\vert t}\! =\! \frac{\partial \Gamma}{\partial \bm x}\! \biggr\vert_{\op}\!\!\!\! \bm{x}_{k+t\vert t} + \frac{\partial \Gamma}{\partial f_{\textnormal{GRF}}}\! \biggr\vert_{\op}\!\!\!\! f_{\textnormal{GRF}, k+t\vert t} + \frac{\partial \Gamma}{\partial \lambda}\! \biggr\vert_{\op}\!\!\!\! \lambda_{k+t\vert t} + \bm{\varepsilon}_{\op},
\end{equation}
where $k=0,1,\cdots, N_\textnormal{h}-1$ and $\bm{x}\in\mathbb{R}^{24}$ represents state variables. Here, $\bm{x}_{k+t\vert t},f_{\textnormal{GRF},k+t\vert t}$, and $\lambda_{k+t\vert t}$ denotes the predicted states, inputs (GRFs), and Lagrange Multiplier at time $k+t$ computed at time $t$, respectively. Additionally, Matrices with subscript `$\op$', and $\bm{\varepsilon}_\op$ is the Jacobian matrices and offset term calculated around $\bm{x}_t$, $f_{\textnormal{GRF},t-1}$, and $\lambda_{t-1}$.
Similar to \eqref{eq:linearizedsrb}, the equality constraint in \eqref{eq:interconnectedSRBdyn} also can be linearized and discretized around the operating point in the following manner:
\begin{equation}\label{eq:linearizedholocon}
    \frac{\partial \ddot{\Lambda}}{\partial \bm x}\! \biggr\vert_{\op}\!\!\!\! \bm{x}_{k+t\vert t} + \frac{\partial \ddot{\Lambda}}{\partial f_{\textnormal{GRF}}}\! \biggr\vert_{\op}\!\!\!\! f_{\textnormal{GRF}, k+t\vert t} + \frac{\partial \ddot{\Lambda}}{\partial \lambda}\! \biggr\vert_{\op}\!\!\!\! \lambda_{k+t\vert t} + \bm{\beta}_{\op}=0.
\end{equation}
The MPC algorithm for cooperative locomotion with holonomically constrained SRB dynamics can thus be written as the optimization problem:
\begin{alignat}{4}\label{eq:srbmpc}
&\min_{(\bm{x},f_{\textnormal{GRF}},\lambda)} &&\|\bm{x}_{t+N|t}-\bm{x}^{\des}_{t+N|t}\|_{\bm{P}}^{2} + \sum_{k=0}^{N_{\textnormal{h}}-1} \|\bm{x}_{k+t|t} - \bm{x}^{\des}_{k+t|t}\|_{\bm{Q}}^{2}\nonumber\\
&&&+ \sum_{k=0}^{N_{\textnormal{h}}-1} \{\|f_{\textnormal{GRF}, k+t|t} \|_{\bm{R}_{f_{\textnormal{GRF}}}}^{2} + \|\lambda_{k+t|t}\|_{\bm{R}_{\lambda}}^{2}\} \nonumber\\
&\quad\quad\textrm{s.t.}&&\textrm{equations \eqref{eq:linearizedsrb}} \textrm{ and }\textrm{\eqref{eq:linearizedholocon}} \textrm{ and } f_\textnormal{GRF} \in \mathcal{FC}
\end{alignat}
where $k=0,1,\cdots,N_{\textnormal{h}}-1$, $\bm{P}$, $\bm{Q}$, and $\bm{R}_{f_{\textnormal{GRF}}}$ are positive definite matrices and $\bm{R}_\lambda$ is positive scalar. The inequality constraints of \eqref{eq:srbmpc} represent the feasibility of GRFs for two agents, where $\mathcal{FC}$ denotes the friction cone condition on each foot contacting the ground. Here we remark that the first element of the optimal state, $x^{\star}_{t+1\vert t}$ and optimal GRFs, $f^{\star}_{\textnormal{GRF},t\vert t}$, computed from \eqref{eq:srbmpc} is applied to the low-level controller for tracking.

\subsection{QP-based Virtual Constraint Controller}
\label{sec:qpvcctrl}
We now low-level distributed controller used to leverage the full-order nonlinear models of individual agents tracking the safety-ensured optimal reduced-order trajectories. Here, we extend the virtual constraints controller of \cite{hamed2020quadrupedal} for the development of distributed controllers for multi-agent systems.
The full-order floating-based model of the $i$th agent can be represented by the Euler-Lagrange equation as follows:
\begin{equation}\label{eq:fullorderdynamics}
    D(q_i)\ddot{q}_i + H(q_i,\dot{q}_i) = B\tau_i+\sum J^{\top}(q_i)f_i,
\end{equation}
where $q_i\in\mathcal{Q}\subset\mathbb{R}^{n_g}$ represents the generalized coordinates of the $i$th robot, $\mathcal{Q}$ and $n_g$ denote the configuration space and the dimension of generalized coordinates, respectively, $\tau_i\in\mathcal{T}\subset\mathbb{R}^{n_j}$ represents the joint-level torques, $\mathcal{T}$ is an admissible torque, and $n_j$ denotes the number of joints. Here, $f_i$ and $J(q_i)$ are each external force on agent $i$ and the Jacobian matrix where each external force is applied on agent $i$, respectively. Moreover, $D(q_i)\in\mathbb{R}^{n_g \times n_g}$ and $H(q_i,\dot{q}_i)\in\mathbb{R}^{n_g}$ represent mass-inertia matrix and Coriolis, centrifugal and gravitational terms, respectively. Additionally, $B\in\mathbb{R}^{n_g\times n_j}$ denotes the input distribution matrix. Here, we note that the distributed low-level controller does not include the dynamic effect of holonomic constraints in the full-order model, as this effect is already taken into consideration by the prescribed optimal trajectories from the safety-critical coordination controller in Section \ref{sec:safetyplanner}. More specifically, the composition of $\sum J^{\top}(q_i)f_i$ in \eqref{eq:fullorderdynamics} only includes the Jacobian matrices of the ground contact points and the GRF on each ground contact point.

Analogous to the distributed whole-body controller in \cite{kim2022layered}, we consider the following time-varying output functions to be regulated for the motion control of the agent $i$ as $y_{i}(\nu_{i},t):=h_{0}(q_{i})-h_{d,i}(t)$, termed \emph{virtual constraints}.
Here, $h_{0}(q_{i})$ represents the set of controlled variables, $h_{d, i}(t)$ denotes the desired evolution of the controlled variables, and $\nu_i = \col(q_i, \dot{q}_i)\in \mathcal{Q}\times \mathbb{R}^{n_g}$ is the full-order state variables. The controlled variables in this paper are chosen as the absolute orientation (i.e., Euler angles) of the $i$th agent together with its Cartesian coordinates of the $i$th agent's COM, and the Cartesian coordinates of the swing feet in the inertial world frame. The desired evolution of the COM position and orientation is generated by the safety-critical trajectory planner in Section \ref{sec:safetyplanner}. Additionally, coordinates of the desired swing foot trajectories are taken as a B\'ezier polynomial starting from the current footholds and ending at the upcoming footholds, computed based on Raibert's heuristic \cite{raibert1989dynamically}. With the dynamics in hand, these virtual constraints together with the nonslippage condition at each stance foot are concatenated into a convex QP to be solved as 1kHz as follows \cite{hamed2020quadrupedal}
\begin{alignat}{4}\label{eq:QPforvc}
&\min_{(\tau_{i},f_{i},\delta_i)} \,\,\,&&\frac{\zeta_1}{2}\|\tau_{i}\|^{2} + \frac{\zeta_2}{2}\|f_{i}-f_{\textnormal{des}, i}\|^{2} + \frac{\zeta_3}{2}\|\delta\|^{2}\\
&\quad\textrm{s.t.} && \!\!\! \ddot{y}(\nu_i, t)=-K_{P}\,y_{i}-\!K_{D}\,\dot{y}_{i}\!+\!\delta_i \,\,\,\textnormal{(Output Dynamics)}\nonumber\\
& &&\!\!\!\ddot{p}_{\textnormal{foot}, i}=0 \qquad \qquad \qquad \qquad \qquad \quad \,\,\, \textnormal{(Nonslippage)}\nonumber\\
& &&\!\!\!\tau_{i}\in\mathcal{T}, \,\,\, f_{i}\in\mathcal{FC}, \qquad \qquad \, \textnormal{(Feasibility conditions)}\nonumber
\end{alignat}
where $\zeta_1$, $\zeta_2$, and $\zeta_3$ are positive weighting factors, and $f_{\textnormal{des},i}$ represents the desired GRFs prescribed from the safety-critical coordination controller. More specifically, $f_{\textnormal{des},i}$ is identical to the first element of the optimal GRFs, $f^{\star}_{\textnormal{GRF}, t\vert t}$, obtained as a result of \eqref{eq:srbmpc}. More detailed  information about the derivation of output dynamics and nonslippage condition in \eqref{eq:QPforvc} can be found in \cite[Appendix A]{kim2022cooperative}. 

\section{Simulation and Experimental Results}\label{sec:validations}
\begin{figure}[t!]
\centering
\includegraphics[draft=false, width=\linewidth]{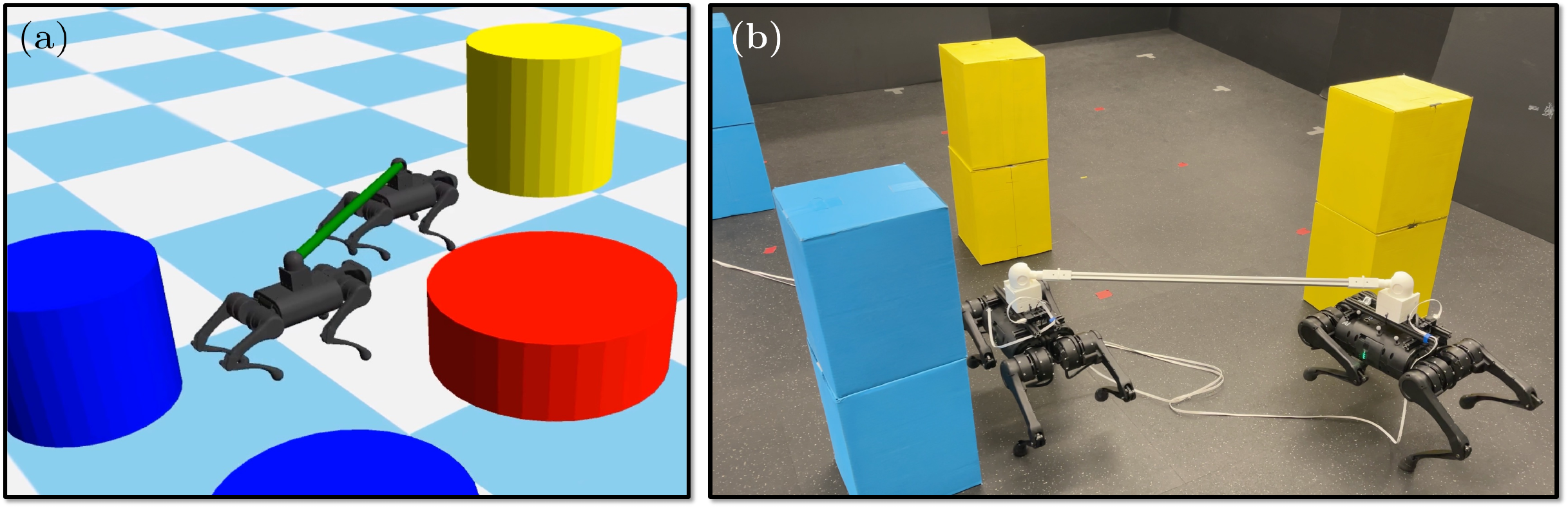}
\vspace{-2.0em}
\caption{Snapshots demonstrating the performance of the proposed three-layered safety-critical coordination controller in (a) full-order simulation experiment and (b) hardware experiment. The left figure shows the snapshot of the simulation experiment with four obstacles in the environment. The right figure shows the snapshot of the hardware experiment with interconnected A1s with four obstacles in the environment.}
\label{fig:pics}
\vspace{-1.5em}
\end{figure}

In this section, we provide the numerical simulations of the safety-critical CBF-based planner and present the simulation and experimental results with the $18$-degrees of freedom (DOFs) quadruped A1 developed by Unitree (see Fig. \ref{fig:pics}). The robot is modeled using a floating base, with the first $6$ DOFs accounting for the unactuated position and orientation of the trunk, while the remaining DOFs correspond to the actuated hip roll, hip pitch, and knee pitch joints for each leg. The robot has a weight of $12.45$ (kg) and the center of the trunk is $0.26$ (m) above the ground during locomotion.
We assume that each agent is holonomically constrained to the other agents, meaning there is no relative translational motion between the agents. To achieve this constraint, we connect the robots together rigidly through a ball joint (refer to Fig. \ref{fig:pics} (b)). In simulations and experiments, this constraint is enforced using a distance constraint. In this section, we use $1$ (m) as the distance constraint between agents. 
Additionally, we use the laptop equipped with an Intel\textsuperscript{\textregistered} Core\textsuperscript{\texttrademark} i7-1185G7 CPU operating at a frequency of $3.00$ GHz with four cores and 16 GB of RAM.

\subsection{Numerical Simulations with Reduced-Order Models}
We study the numerical simulations with interconnected reduced-order models for validating the effectiveness of the safety-critical planner.
We tested two, three, and four obstacles ($D\in \{2,3,4\}$) in the environment with two holonomically constrained reduced-order models ($N=2$). 
The positions of the obstacles are $\col(1.5, \, 0.7)$ (m) and $\col(1.5, \, -0.5)$ (m) for two obstacles with $0.3$ (m) radius for both. One additional obstacle at $\col(2.5, \, 0.1)$ (m) with $0.5$ (m) radius is added to the two obstacle environments for three obstacles. Similarly, The other additional obstacle at $\col(3.5, \, 1.3)$ (m) with $0.4$ (m) radius is added to the three obstacle environments for four obstacles.
The total number of spatial positions between agents for applying CBFs is four ($\eta = 4$). The initial positions of the agent $1$ and $2$ are $\col (0,\,0)$ (m) and $\col (0,\,-1)$ (m), respectively. The expected goal position for the agent $1$ and $2$ are $\col(4,\,0.5)$ (m) and $\col(4,\,-0.5)$ (m), respectively. $k_d(\bm{\varphi})$ is $0.1\{\col(4,\,0.5,\,4,\,-0.5)\!-\!\bm{\varphi}\}\in\mathbb{R}^4$ and 
$A$ is $\diag(0.1\mathbb{I},\, 0.001, 0.001)$ where $\mathbb{I}$ is $12\!\times \!12$, $18\!\times \!18$, and $24\!\times \!24$ identity matrix when two, three, and four obstacles are in the environment, respectively.
In CBFs for holonomic constraint, $\epsilon\!=\!\!5\!\!\times \!\!10^{-5}$ for ensuring the rigid distance constraint between agents. The $\bm{P}_s$ that we used is $0.5\mathbb{I}_4$.
The proposed safety-critical planner is solved using qpSWIFT \cite{pandala2019qpswift}. The solving time of the QP-formulated safety-critical planner takes approximately $0.9$ (ms), $1.2$ (ms), and $1.3$ (ms) for two, three, and four obstacles, respectively.

\begin{figure}[t!]
\centering
\includegraphics[draft=false, width=\linewidth]{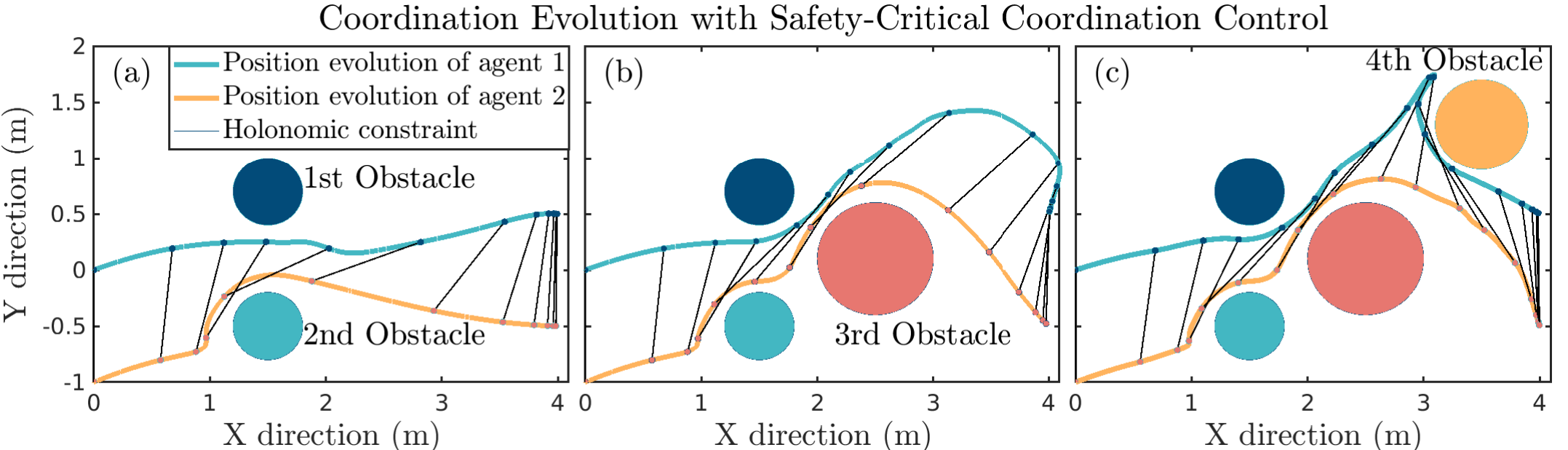}
\vspace{-2.0em}
\caption{Plots of the optimal trajectories for safe cooperative locomotion with (a) two obstacles, (b) three obstacles, and (c) four obstacles in front of the holonomically constrained agents. Black lines show the coordination of the agents by describing the planar position and orientation of the holonomic constraint during the evolution.}
\label{fig:numericalsim}
\vspace{-1.0em}
\end{figure}

\begin{figure}[t!]
\centering
\includegraphics[draft=false, width=\linewidth]{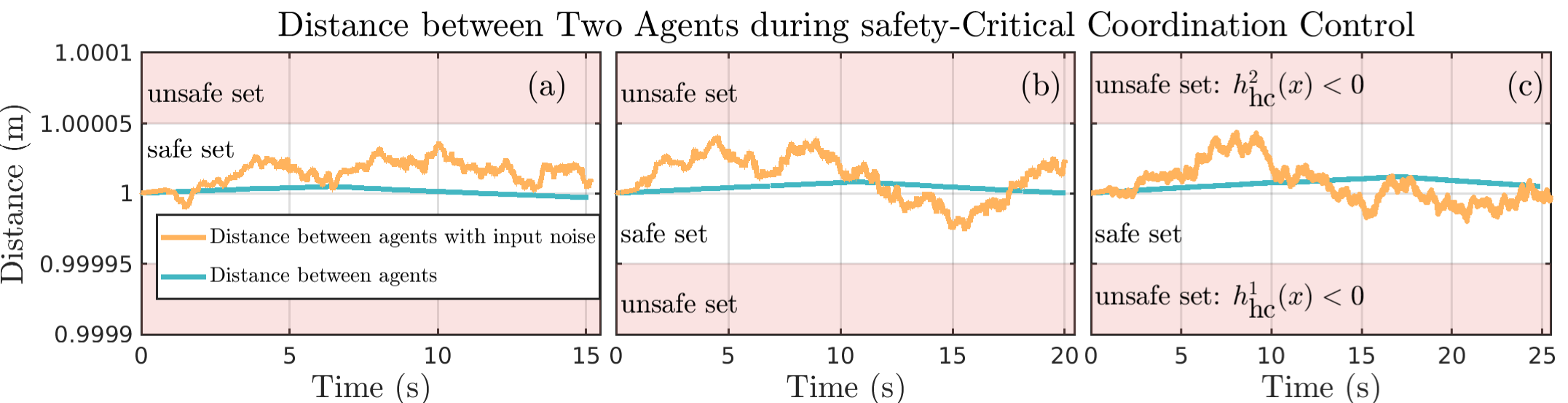}
\vspace{-2.0em}
\caption{Plots of the distance between agents during the position evolution with (a) two obstacles, (b) three obstacles, and (c) four obstacles with and without input noise.}
\label{fig:numericalissf}
\vspace{-2.0em}
\end{figure}

The position evolution of each agent during safety-critical coordination control is depicted in Fig. \ref{fig:numericalsim}. The black lines represent the position and orientation of the holonomic constraint between agents during the evolution. The plot shows that the proposed safety-criticial planner addresses safe cooperation while avoiding two, three, and four obstacles in Figs. \ref{fig:numericalsim} (a), (b), and (c), respectively. The proposed safety-critical planner addresses the obstacle evasion by changing the orientation of the holonomic constraint during evolution. The distance between the agent during position evolution while avoiding two, three, and four obstacles are shown in Figs. \ref{fig:numericalissf} (a), (b), and (c), respectively. With the control input without the noise, we observe that the distance between two agents is close to the distance constraint that we imposed at the beginning. As we choose $\epsilon=5\!\!\times \!\!10^{-5}$, the distance between the agents still stays in the safe set when the control input subject to noise (-43 (dB)) is applied.

The effectiveness of the proposed safety-critical planner for enabling safe cooperative locomotion with holonomic constraint while avoiding obstacles is demonstrated in the numerical simulations presented in Figs. \ref{fig:numericalsim} and \ref{fig:numericalissf}. These numerical simulations illustrate how the proposed framework based on CBFs provides a robustness against control input noise, which is supported by the ISSf CBFs utilized (see Section \ref{sec:issf}). Notably, the framework's robustness can be further enhanced by adjusting the $\epsilon$ parameter in \eqref{eq:issfcbfforholocon} to suit specific implementation requirements. We remark that these results demonstrate the potential of the proposed framework to enable safe and robust cooperative locomotion in various real-world scenarios.

\begin{figure}[t!]
\centering
\includegraphics[draft=false, width=\linewidth]{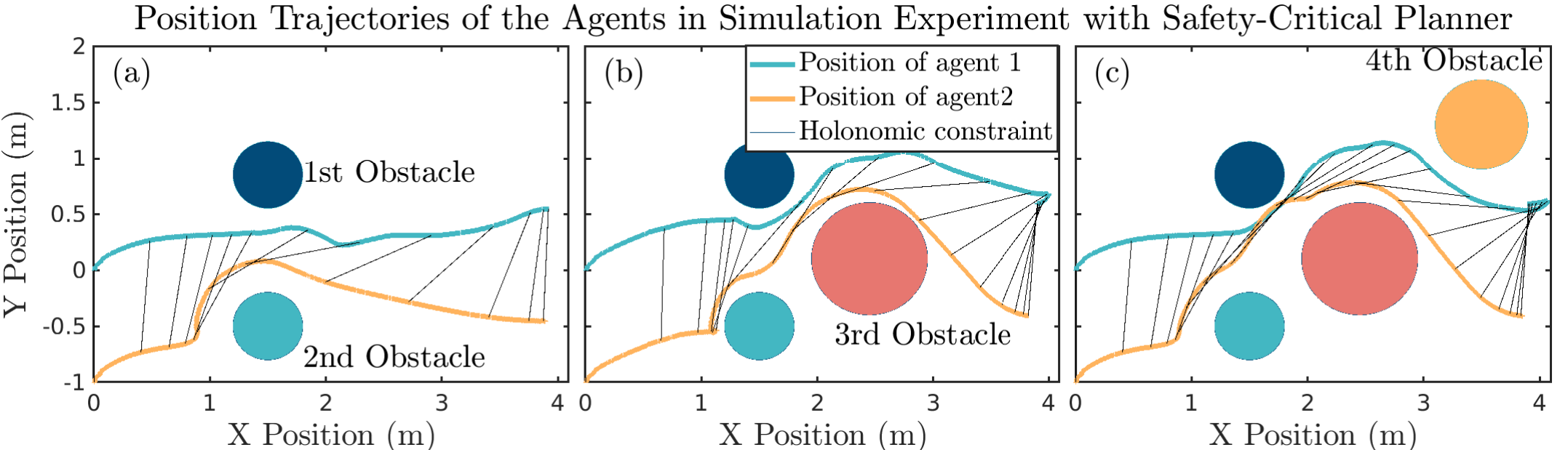}
\vspace{-2.0em}
\caption{Plots of the actual trajectories  of the interconnected full-order model in simulation experiment with (a) two obstacles, (b) three obstacles, and (c) four obstacles. Black lines show the coordination of the agents by describing the planar position and orientation of the holonomic constraint during the evolution.}
\label{fig:fullordersim}
\vspace{-1.0em}
\end{figure}

\begin{figure}[t!]
\centering
\includegraphics[draft=false, width=\linewidth]{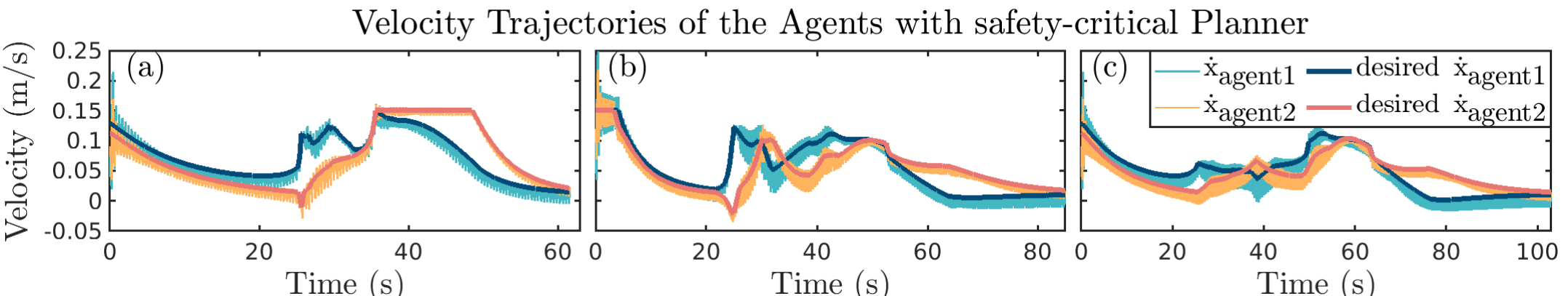}
\vspace{-2.0em}
\caption{Plots of the safety-ensured optimal velocities and actual velocities of the interconnected full-order model in simulation experiment with (a) two obstacles, (b) three obstacles, and (c) four obstacles.}
\label{fig:fullordersimvel}
\vspace{-1.5em}
\end{figure}

\subsection{Experimental Validations}
We next validate the effectiveness of the safety-critical planner with full-order simulation experiments (see Fig. \ref{fig:pics} (a)) and hardware experiments (see Fig. \ref{fig:pics} (b)). We used RaiSim \cite{raisimpaper} for the simulation environment. The same safety-critical planner validated in numerical simulations is implemented as the top-level planner to ensure safe cooperative locomotion. The top-level safety-critical planner runs in $200$ (Hz).
In the middle level of the trajectory planner, MPC with interconnected SRB dynamics runs in $200$ (Hz). 
The control horizon for the MPC is taken as $N_h=6$ discrete-time samples. MPC is solved using qpSWIFT \cite{pandala2019qpswift} and takes approximately $1.6$ (ms) with $294$ decision variables. The stage cost gain of the MPC is tuned as $\bm{Q} = \diag\{Q_{\phi_{\textnormal{COM},1}} \ Q_{\dot{\phi}_{\textnormal{COM},1}} \ Q_{\xi 1} \ Q_{\omega1} \ Q_{\phi_{\textnormal{COM},2}} \ Q_{\dot{\phi}_{\textnormal{COM},2}} \ Q_{\xi 2} \ Q_{\omega2} \} \in \Real^{24\times 24}$, where $\bm{Q}_{\phi_{\textnormal{COM},i}} = 10^{5}\times\diag\{3 \ 300 \ 30\}$, $\bm{Q}_{\dot{\phi}_{\textnormal{COM},1}} = 10^4\,\iden_{3\times 3}$, $\bm{Q}_{\xi i} = 10^8\,\iden_{3\times 3}$, and $\bm{Q}_{\omega i} = 5\times 10^3\,\iden_{3\times 3}$, $i\in\mathcal{I}$. The terminal cost gain of the MPC is also tuned as $\bm{P} = 10^{-1} \bm{Q} \in \Real^{24\times 24}$. The input gains of the MPC are chosen as $\bm{R}_{f_{\textnormal{GRF}}} = 10^{-2}\,\iden_{24\times 24}$ and $\bm{R}_{\lambda} = 10^{4}$. In the full-order simulation experiment, each obstacle is kept at a consistent dimension and position, while in the hardware experiment, the dimension and position of each obstacle are determined based on the environment measurement. More specifically, while the arrangement of the obstacles in the hardware experiments was similar to that used in simulations, the detailed dimensions and positions differed between the simulation and hardware experiments.

\begin{figure}[t!]
\centering
\includegraphics[draft=false, width=\linewidth]{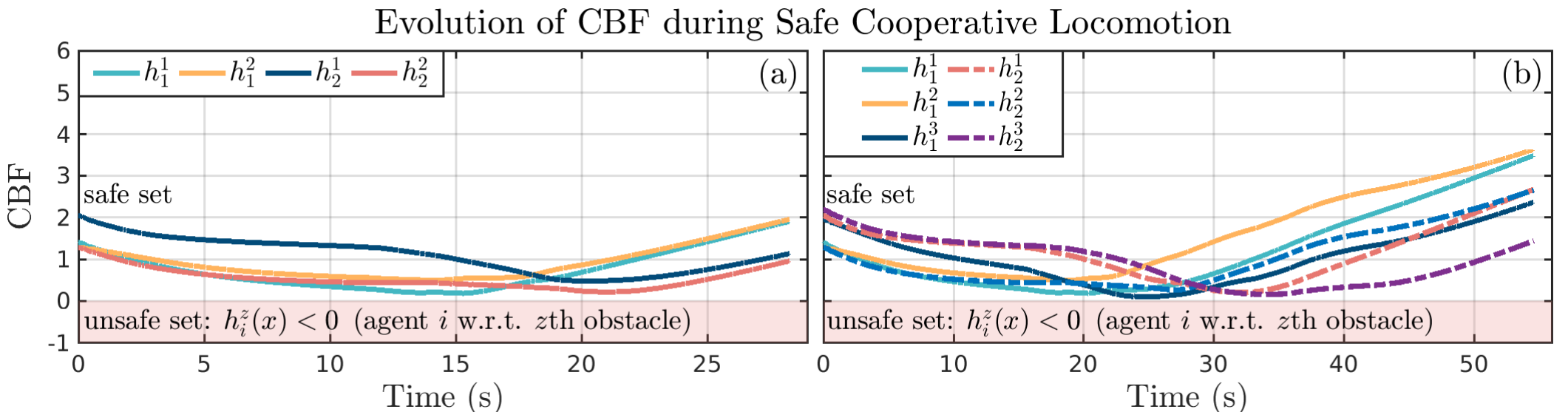}
\vspace{-2.0em}
\caption{Evolution of the CBFs on each agent with respect to (a) two obstacles and (b) three obstacles in hardware experiments. The red-colored area represents the unsafe set, defined with the sub-level set of the CBFs.}
\label{fig:cbfsforexp}
\vspace{-1.5em}
\end{figure}

The position and velocity evolutions of the interconnected full-order models in simulation experiments are described in Fig. \ref{fig:fullordersim} and \ref{fig:fullordersimvel}, respectively. The plot shows the proposed layered safety-critical controller addresses safe cooperative locomotion with holonomically constrained full-order models subject to two, three, and four obstacles in Figs. \ref{fig:fullordersim} (a), (b), and (c), respectively. The safety-ensured desired velocities and the evolution of actual velocities are depicted in Fig. \ref{fig:fullordersimvel}. The plot shows that the proposed safety-critical coordination controller generates the safety-ensured trajectories and enables the holonomically constrained full-order systems to follow those trajectories without losing stability.

We next investigate the performance of the closed-loop system with the safety-critical coordination controller with hardware, as shown in Figs. \ref{fig:titlepic} and \ref{fig:pics} (b). These experiments include two and three obstacles to be avoided. The evolution of the CBFs of each agent subject to individual obstacles are shown in Fig. \ref{fig:cbfsforexp}. As the CBFs stay in the safe set area, we can conclude that the cooperative locomotion of the holonomically constrained full-order system successfully tracks the safety-ensured trajectories. Videos of all experiments are available online \cite{YouTube_cbfcooperative}.

\section{Conclusion}\label{sec:conclusion}
This work has demonstrated the potential of using CBFs to achieve safe cooperative locomotion of holonomically constrained quadrupedal robots. Leveraging ISSf CBFs in the interconnected reduced-order model, we developed a layered controller with the safety-critical planner that successfully accounted for the safety of complex dynamics during locomotion. Our simulations and hardware experiments showed the effectiveness of the proposed method in achieving safety-ensured cooperative locomotion subject to holonomic constraints. In future work, we will explore how this framework could extend to an even greater number of agents in a more complex and uncertain environment. This includes the extension of the safety-critical planner for cooperation to three dimensions to allow for varying terrain types.




\bibliographystyle{IEEEtran}
\bibliography{references}

\begin{thebibliography}{10}
\providecommand{\url}[1]{#1}
\csname url@samestyle\endcsname
\providecommand{\newblock}{\relax}
\providecommand{\bibinfo}[2]{#2}
\providecommand{\BIBentrySTDinterwordspacing}{\spaceskip=0pt\relax}
\providecommand{\BIBentryALTinterwordstretchfactor}{4}
\providecommand{\BIBentryALTinterwordspacing}{\spaceskip=\fontdimen2\font plus
\BIBentryALTinterwordstretchfactor\fontdimen3\font minus
  \fontdimen4\font\relax}
\providecommand{\BIBforeignlanguage}[2]{{%
\expandafter\ifx\csname l@#1\endcsname\relax
\typeout{** WARNING: IEEEtran.bst: No hyphenation pattern has been}%
\typeout{** loaded for the language `#1'. Using the pattern for}%
\typeout{** the default language instead.}%
\else
\language=\csname l@#1\endcsname
\fi
#2}}
\providecommand{\BIBdecl}{\relax}
\BIBdecl

\bibitem{culbertson2021decentralized}
P.~Culbertson, J.-J. Slotine, and M.~Schwager, ``Decentralized adaptive control
  for collaborative manipulation of rigid bodies,'' \emph{IEEE Transactions on
  Robotics}, vol.~37, no.~6, pp. 1906--1920, 2021.

\bibitem{machado2016multi}
T.~Machado, T.~Malheiro, S.~Monteiro, W.~Erlhagen, and E.~Bicho,
  ``Multi-constrained joint transportation tasks by teams of autonomous mobile
  robots using a dynamical systems approach,'' in \emph{IEEE International
  Conference on Robotics and Automation}, 2016, pp. 3111--3117.

\bibitem{tagliabue2019robust}
A.~Tagliabue, M.~Kamel, R.~Siegwart, and J.~Nieto, ``Robust collaborative
  object transportation using multiple {MAVs},'' \emph{The International
  Journal of Robotics Research}, vol.~38, no.~9, pp. 1020--1044, 2019.

\bibitem{wehbeh2020distributed}
J.~Wehbeh, S.~Rahman, and I.~Sharf, ``Distributed model predictive control for
  {UAVs} collaborative payload transport,'' in \emph{IEEE/RSJ International
  Conference on Intelligent Robots and Systems}, 2020, pp. 11\,666--11\,672.

\bibitem{kim2022cooperative}
J.~Kim and K.~Akbari~Hamed, ``Cooperative locomotion via supervisory predictive
  control and distributed nonlinear controllers,'' \emph{Journal of Dynamic
  Systems, Measurement, and Control}, vol. 144, no.~3, 2022.

\bibitem{kim2022layered}
J.~Kim, R.~T. Fawcett, V.~R. Kamidi, A.~D. Ames, and K.~A. Hamed, ``Layered
  control for cooperative locomotion of two quadrupedal robots: Centralized and
  distributed approaches,'' \emph{arXiv preprint arXiv:2211.06913}, 2022.

\bibitem{fawcett2022distributed}
R.~T. Fawcett, L.~Amanzadeh, J.~Kim, A.~D. Ames, and K.~A. Hamed, ``Distributed
  data-driven predictive control for multi-agent collaborative legged
  locomotion,'' \emph{arXiv preprint arXiv:2211.06917}, 2022.

\bibitem{ames2016control}
A.~D. Ames, X.~Xu, J.~W. Grizzle, and P.~Tabuada, ``Control barrier function
  based quadratic programs for safety critical systems,'' \emph{IEEE
  Transactions on Automatic Control}, vol.~62, no.~8, pp. 3861--3876, 2016.

\bibitem{ames2019control}
A.~D. Ames, S.~Coogan, M.~Egerstedt, G.~Notomista, K.~Sreenath, and P.~Tabuada,
  ``Control barrier functions: Theory and applications,'' in \emph{2019 18th
  European control conference (ECC)}.\hskip 1em plus 0.5em minus 0.4em\relax
  IEEE, 2019, pp. 3420--3431.

\bibitem{wang2017safety}
L.~Wang, A.~D. Ames, and M.~Egerstedt, ``Safety barrier certificates for
  collisions-free multirobot systems,'' \emph{IEEE Transactions on Robotics},
  vol.~33, no.~3, pp. 661--674, 2017.

\bibitem{zhao2017defend}
S.~Zhao and Z.~Sun, ``Defend the practicality of single-integrator models in
  multi-robot coordination control,'' in \emph{2017 13th IEEE International
  Conference on Control \& Automation (ICCA)}.\hskip 1em plus 0.5em minus
  0.4em\relax IEEE, 2017, pp. 666--671.

\bibitem{pickem2017robotarium}
D.~Pickem, P.~Glotfelter, L.~Wang, M.~Mote, A.~Ames, E.~Feron, and
  M.~Egerstedt, ``The robotarium: A remotely accessible swarm robotics research
  testbed,'' in \emph{2017 IEEE International Conference on Robotics and
  Automation (ICRA)}.\hskip 1em plus 0.5em minus 0.4em\relax IEEE, 2017, pp.
  1699--1706.

\bibitem{chen2020guaranteed}
Y.~Chen, A.~Singletary, and A.~D. Ames, ``Guaranteed obstacle avoidance for
  multi-robot operations with limited actuation: A control barrier function
  approach,'' \emph{IEEE Control Systems Letters}, vol.~5, no.~1, pp. 127--132,
  2020.

\bibitem{squires2021model}
E.~Squires, R.~Konda, S.~Coogan, and M.~Egerstedt, ``Model free barrier
  functions via implicit evading maneuvers,'' \emph{arXiv preprint
  arXiv:2107.12871}, 2021.

\bibitem{molnar2021model}
T.~G. Molnar, R.~K. Cosner, A.~W. Singletary, W.~Ubellacker, and A.~D. Ames,
  ``Model-free safety-critical control for robotic systems,'' \emph{IEEE
  robotics and automation letters}, vol.~7, no.~2, pp. 944--951, 2021.

\bibitem{singh2020robust}
S.~Singh, M.~Chen, S.~L. Herbert, C.~J. Tomlin, and M.~Pavone, ``Robust
  tracking with model mismatch for fast and safe planning: an sos optimization
  approach,'' in \emph{Algorithmic Foundations of Robotics XIII: Proceedings of
  the 13th Workshop on the Algorithmic Foundations of Robotics 13}.\hskip 1em
  plus 0.5em minus 0.4em\relax Springer, 2020, pp. 545--564.

\bibitem{grandia2021multi}
R.~Grandia, A.~J. Taylor, A.~D. Ames, and M.~Hutter, ``Multi-layered safety for
  legged robots via control barrier functions and model predictive control,''
  in \emph{2021 IEEE International Conference on Robotics and Automation
  (ICRA)}.\hskip 1em plus 0.5em minus 0.4em\relax IEEE, 2021, pp. 8352--8358.

\bibitem{khazoom2022humanoid}
C.~Khazoom, D.~Gonzalez-Diaz, Y.~Ding, and S.~Kim, ``Humanoid self-collision
  avoidance using whole-body control with control barrier functions,'' in
  \emph{2022 IEEE-RAS 21st International Conference on Humanoid Robots
  (Humanoids)}.\hskip 1em plus 0.5em minus 0.4em\relax IEEE, 2022, pp.
  558--565.

\bibitem{sontag1999control}
E.~D. Sontag, ``Control-lyapunov functions,'' \emph{Open problems in
  mathematical systems and control theory}, pp. 211--216, 1999.

\bibitem{khalil2002nonlinear}
H.~Khalil, \emph{Nonlinear systems}, 3rd~ed.\hskip 1em plus 0.5em minus
  0.4em\relax Prentice Hall, 2002.

\bibitem{kolathaya2018input}
S.~Kolathaya and A.~D. Ames, ``Input-to-state safety with control barrier
  functions,'' \emph{IEEE control systems letters}, vol.~3, no.~1, pp.
  108--113, 2018.

\bibitem{ding2021representation}
Y.~Ding, A.~Pandala, C.~Li, Y.-H. Shin, and H.-W. Park, ``Representation-free
  model predictive control for dynamic motions in quadrupeds,'' \emph{IEEE
  Transactions on Robotics}, vol.~37, no.~4, pp. 1154--1171, 2021.

\bibitem{chignoli2020variational}
M.~Chignoli and P.~M. Wensing, ``Variational-based optimal control of
  underactuated balancing for dynamic quadrupeds,'' \emph{IEEE Access}, vol.~8,
  pp. 49\,785--49\,797, 2020.

\bibitem{hamed2020quadrupedal}
K.~A. Hamed, J.~Kim, and A.~Pandala, ``Quadrupedal locomotion via event-based
  predictive control and qp-based virtual constraints,'' \emph{IEEE Robotics
  and Automation Letters}, vol.~5, no.~3, pp. 4463--4470, 2020.

\bibitem{raibert1989dynamically}
M.~H. Raibert, H.~B. Brown~Jr, M.~Chepponis, J.~Koechling, and J.~K. Hodgins,
  ``Dynamically stable legged locomotion,'' Massachusetts Inst of Tech
  Cambridge Artificial Intelligence Lab, Tech. Rep., 1989.

\bibitem{pandala2019qpswift}
A.~G. Pandala, Y.~Ding, and H.-W. Park, ``qpswift: A real-time sparse quadratic
  program solver for robotic applications,'' \emph{IEEE Robotics and Automation
  Letters}, vol.~4, no.~4, pp. 3355--3362, 2019.

\bibitem{raisimpaper}
J.~Hwangbo, J.~Lee, and M.~Hutter, ``Per-contact iteration method for solving
  contact dynamics,'' \emph{IEEE Robotics and Automation Letters}, vol.~3,
  no.~2, pp. 895--902, 2018.

\bibitem{YouTube_cbfcooperative}
Safety-critical coordination for cooperative legged locomotion via control
  barrier functions. {[Online]. Available}:
  \url{https://vimeo.com/803721560/bd1871402a}.

\end{thebibliography}

\end{document}